\def\year{2018}\relax
\documentclass[letterpaper]{article} 
\usepackage{aaai18}  
\usepackage{times}  
\usepackage{helvet}  
\usepackage{courier}  
\usepackage{url}  
\usepackage{graphicx}  
\usepackage{amsmath}
\usepackage{amssymb}
\usepackage{algorithm}
\usepackage{algorithmic}
\usepackage{tabularx}
\usepackage{booktabs}
\usepackage{multirow}
\usepackage{subcaption}
\usepackage{pdfpages}

\begin{document}
%
\title{Knowledge Graph Embedding with Iterative Guidance from Soft Rules}
\author{Shu Guo$^{1,2}$, Quan Wang$^{1,2,3}$\thanks{Corresponding author: Quan Wang (wangquan@iie.ac.cn).}, Lihong Wang$^{4}$, Bin Wang$^{1,2}$, Li Guo$^{1,2}$ \\
  $^{1}$Institute of Information Engineering, Chinese Academy of Sciences\\
  $^{2}$School of Cyber Security, University of Chinese Academy of Sciences\\
  $^{3}$State Key Laboratory of Information Security, Chinese Academy of Sciences\\
  $^{4}$National Computer Network Emergency Response Technical Team \& Coordination Center of China}
\maketitle
\begin{abstract}
  Embedding knowledge graphs (KGs) into continuous vector spaces is a focus of current research. Combining such an embedding model with logic rules has recently attracted increasing attention. Most previous attempts made a one-time injection of logic rules, ignoring the interactive nature between embedding learning and logical inference. And they focused only on hard rules, which always hold with no exception and usually require extensive manual effort to create or validate. In this paper, we propose Rule-Guided Embedding (RUGE), a novel paradigm of KG embedding with iterative guidance from soft rules. RUGE enables an embedding model to learn simultaneously from 1) labeled triples that have been directly observed in a given KG, 2) unlabeled triples whose labels are going to be predicted iteratively, and 3) soft rules with various confidence levels extracted automatically from the KG. In the learning process, RUGE iteratively queries rules to obtain soft labels for unlabeled triples, and integrates such newly labeled triples to update the embedding model. Through this iterative procedure, knowledge embodied in logic rules may be better transferred into the learned embeddings. We evaluate RUGE in link prediction on Freebase and YAGO. Experimental results show that: 1) with rule knowledge injected iteratively, RUGE achieves significant and consistent improvements over state-of-the-art baselines; and 2) despite their uncertainties, automatically extracted soft rules are highly beneficial to KG embedding, even those with moderate confidence levels. The code and data used for this paper can be obtained from \url{https://github.com/iieir-km/RUGE}.
\end{abstract}

\section{Introduction}
Knowledge graphs (KGs) such as WordNet~\cite{miller1995:WordNet}, Freebase~\cite{bollacker2008:FreeBase}, YAGO~\cite{suchanek2007:YAGO}, and NELL~\cite{carlson2010:NELL} are extremely useful resources for many AI related applications. A KG is a multi-relational graph composed of entities as nodes and relations as different types of edges. Each edge is represented as a triple (\textit{head entity}, \textit{relation}, \textit{tail entity}), indicating that there is a specific relation between two entities, e.g., (\texttt{\small Paris}, \texttt{\small CapitalOf}, \texttt{\small France}). Although effective in representing structured data, the underlying symbolic nature of such triples often makes KGs hard to manipulate.

Recently, a new research direction termed as \textit{knowledge graph embedding} has been proposed and quickly received massive attention~\cite{nickel2011:RESCAL,bordes2013:TransE,wang2014:TransH,lin2015:TransR,yang2015:DistMult,nickel2016:HolE,trouillon2016:ComplEx}. The key idea is to embed entities and relations in a KG into a low-dimensional continuous vector space, so as to simplify the manipulation while preserving the inherent structure of the KG. Such embeddings contain rich semantic information, and can benefit a broad range of downstream applications~\cite{weston2013:RE,bordes2014:SME,zhang2016:RecSys,xiong2017:ESR}.

Traditional methods performed embedding based solely on triples observed in a KG. But considering the power of logic rules in knowledge acquisition and inference, combining embedding models with logic rules has become a focus of current research~\cite{rocktaschel2014:EmbedLogic,vendrov2015:OrderEmbeddings,wang2016:LogicEmbeddings,hu2016:RuleCNN}. Wang et al. \shortcite{wang2015:ERInfer} and Wei et al.~\shortcite{wei2015:EmbedMLN} tried to use embedding models and logic rules for KG completion. But in their work, rules are modeled separately from embedding models, and would not help to learn more predictive embeddings. Rockt\"{a}schel et al. \shortcite{rocktaschel2015:EmbedLogic} and Guo et al.~\shortcite{guo2016:KALE} then devised joint learning paradigms which can inject first-order logic (FOL) into KG embedding. Demeester et al.~\shortcite{demeester2016:LiftedRule} further proposed lifted rule injection to avoid the costly propositionalization of FOL rules. Although these joint models are able to learn better embeddings after integrating logic rules, they still have their drawbacks and restrictions.

First of all, these joint models made a one-time injection of logic rules, taking them as additional rule-based training instances~\cite{rocktaschel2015:EmbedLogic} or regularization terms~\cite{demeester2016:LiftedRule}. We argue that rules can better enhance KG embedding, however, in an iterative manner. Given the learned embeddings and their rough predictions, rules can be used to refine the predictions and infer new facts. The newly inferred facts, in turn, will help to learn better embeddings and more accurate logical inference. Previous methods fail to model such interactions between embedding models and logic rules. Furthermore, they focused only on hard rules which always hold with no exception. Such rules usually require extensive manual effort to create or validate. Actually, besides hard rules, a significant amount of background information can be encoded as soft rules, e.g., ``a person is very likely (but not necessarily) to have a nationality of the country where he/she was born''. Soft rules can be extracted automatically and efficiently via modern rule mining systems~\cite{galarraga2013:AMIE,galarraga2015:AMIE+}. Yet, despite this merit, soft rules have not been well studied in previous methods.

\begin{figure}[t]
\centering
  \includegraphics[width=0.4\textwidth]{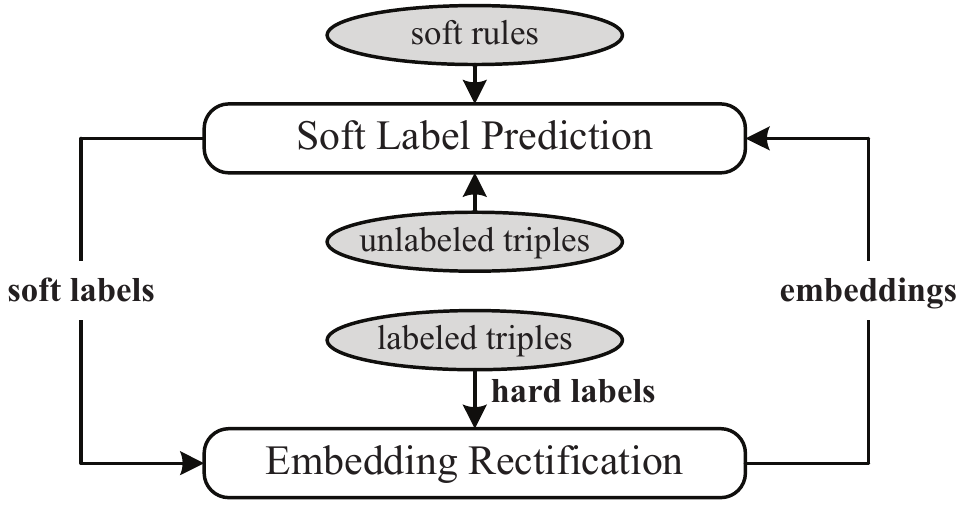}
  \caption{Framework overview. RUGE enables an embedding model to learn simultaneously from labeled triples, unlabeled triples, and soft rules in an iterative manner, where each iteration alternates between a soft label prediction stage and an embedding rectification stage.}\label{fig:RUGE}
\end{figure}

This paper proposes \textit{RUle-Guided Embedding} (RUGE), a novel paradigm of KG embedding with iterative guidance from soft rules. As sketched in Fig.~\ref{fig:RUGE}, it enables an embedding model to learn simultaneously from 1) labeled triples that have been directly observed in a given KG, 2) unlabeled triples whose labels are going to be predicted iteratively, and 3) soft rules with different confidence levels extracted automatically from the KG. During each iteration of the learning process, the model alternates between a soft label prediction stage and an embedding rectification stage. The former uses currently learned embeddings and soft rules to predict soft labels for unlabeled triples, and the latter further integrates both labeled and unlabeled triples (with hard and soft labels respectively) to update current embeddings. Through this iterative procedure, knowledge embodied in logic rules may be better transferred into the learned embeddings.

We empirically evaluate RUGE on large scale public KGs, namely Freebase and YAGO. Experimental results reveal that: 1) by incorporating logic rules, RUGE significantly and consistently improves over state-of-the-art basic embedding models (without rules); 2) compared to those one-time injection schemes studied before, the iterative injection strategy maximizes the utility of logic rules for KG embedding, and indeed achieves substantially better performance; 3) despite the uncertainties, automatically extracted soft rules are highly beneficial to KG embedding, even those with moderate confidence levels.

The contributions of this paper are threefold. 1) We devise a novel paradigm of KG embedding which iteratively injects logic rules into the learned embeddings. To our knowledge, this is the first work that models interactions between embedding learning and logical inference in a principled framework. 2) We demonstrate the usefulness of automatically extracted soft rules in KG embedding, thereby eliminating the requirement of laborious manual rule creation. 3) Our approach is quite generic and flexible. It can integrate various types of rules with different confidence levels to enhance a good variety of KG embedding models.

\section{Related Work}
Recent years have witnessed increasing interest in learning distributed representations for entities and relations in KGs, a.k.a. KG embedding. Various techniques have been devised for this task, e.g., translation-based models which take relations as translating operations between head and tail entities~\cite{bordes2013:TransE,wang2014:TransH,lin2015:TransR}, simple compositional models which match compositions of head-tail entity pairs with their relations~\cite{nickel2011:RESCAL,yang2015:DistMult,nickel2016:HolE,trouillon2016:ComplEx}, and neural networks which further introduce non-linear layers and deep architectures \cite{socher2013:NTN,bordes2014:SME,dong2014:KnowledgeVault,liu2016:NAM}. Among these techniques, ComplEx~\cite{trouillon2016:ComplEx}, a compositional model which represents entities and relations as complex-valued vectors, achieves a very good trade-off between accuracy and efficiency. Most of the currently available techniques perform the embedding task based solely on triples observed in a KG. Some recent work further tried to use other information, e.g., entity types \cite{guo2015:SSE,xie2016:TKRL} and textual descriptions \mbox{\cite{xie2016:DKRL,xiao2017:SSP}}, to learn more predictive embeddings. See~\cite{wang2017:KGEmbedding} for a thorough review of KG embedding techniques.

Given the power of logic rules in knowledge acquisition and inference, combining KG embedding with logic rules becomes a focus of current research. Wang et al.~\shortcite{wang2015:ERInfer} and Wei et al.~\shortcite{wei2015:EmbedMLN} devised pipelined frameworks which use logic rules to further refine predictions made by embedding models. In their work, rules will not help to learn better embeddings. Rockt\"{a}schel et al.~\shortcite{rocktaschel2015:EmbedLogic} and Guo et al.~\shortcite{guo2016:KALE} then tried to learn KG embeddings jointly from triples and propositionalized FOL rules. Demeester et al.~\shortcite{demeester2016:LiftedRule} further proposed lifted rule injection to avoid the costly propositionalization. These joint models, however, made a one-time injection of logic rules, ignoring the interactive nature between embedding learning and logical inference. Moreover, they can only handle hard rules which are usually manually created or validated.

Besides logic rules, relation paths which can be regarded as Horn clauses and get a strong connection to logical inference~\cite{gardner2015:Paths}, have also been studied in KG embedding~\cite{neelakantan2015:CompositionalVSM,lin2015:PTransE,guu2015:TraverseKG}. But in these methods, relation paths are incorporated, again, in a one-time manner. Our approach, in contrast, iteratively injects knowledge contained in logic rules into KG embedding, and is able to handle soft rules with various confidence levels extracted automatically from KGs.

Combining logic rules with distributed representations is also an active research topic in other contexts outside KGs. Faruqui et al.~\shortcite{faruqui2014:WordEmbedding} tried to inject ontological knowledge from WordNet into word embeddings. Vendrov et al.~\shortcite{vendrov2015:OrderEmbeddings} introduced order-embedding to model the partial order structure of hypernymy, textual entailment, and image captioning. Hu et al.~\shortcite{hu2016:RuleCNN} proposed to enhance various types of neural networks with FOL rules. All these studies demonstrate the capability of logic rules to enhance distributed representation learning.

\section{Rule-Guided Knowledge Graph Embedding}
This section introduces \textit{RUle-Guided Embedding} (RUGE), a novel paradigm of KG embedding with iterative guidance from soft rules. RUGE enables an embedding model to learn simultaneously from labeled triples, unlabeled triples, and soft rules in an iterative manner. During each iteration, the model alternates between a soft label prediction stage and an embedding rectification stage. Fig.~\ref{fig:RUGE} sketches this overall framework. In what follows, we first describe our learning resources, and then detail the two alternating stages.

\subsection{Learning Resources}
Suppose we are given a KG with a set of triples observed, i.e., $\mathcal{O} = \{ (e_i, r_k, e_j) \}$. Each triple is composed of two entities $e_i, e_j \in \mathcal{E}$ and their relation $r_k \in \mathcal{R}$, where $\mathcal{E}$ and $\mathcal{R}$ are the sets of entities and relations respectively. We obtain our learning resources (i.e., labeled triples, unlabeled triples, and soft rules) and model them as follows.

\smallskip
\noindent\textbf{Labeled Triples.} We take the triples observed in $\mathcal{O}$ as positive ones. For each positive triple $(e_i, r_k, e_j)$, we randomly corrupt the head $e_i$ or the tail $e_j$, to form a negative triple $(e_i', r_k, e_j)$ or $(e_i, r_k, e_j')$, where $e_i'\in\mathcal{E}\setminus\{e_i\}$ and $e_j'\in\mathcal{E}\setminus\{e_j\}$. We denote a labeled triple as $x_\ell$, and associate with it a label $y_\ell=1$ if $x_\ell$ is positive, and $y_\ell=0$ otherwise. Let $\mathcal{L}=\{(x_\ell, y_\ell)\}$ denote the set of these labeled triples (along with their labels).

\smallskip
\noindent\textbf{Unlabeled Triples.} Besides the labeled triples, we collect a set of unlabeled triples $\mathcal{U}=\{x_u\}$, where $x_u=(e_i,r_k,e_j)$ indicates an unlabeled triple. In fact, all the triples that have not been observed in $\mathcal{O}$ can be taken as unlabeled ones. But in this paper, we consider only those encoded in the conclusion of a soft rule, as detailed below.

\smallskip
\noindent\textbf{Soft Rules.} We also consider a set of FOL rules with different confidence levels, denoted as $\mathcal{F}=\{(f_p,\lambda_p)\}_{p=1}^{P}$. Here, $f_p$ is the $p$-th logic rule defined over the given KG, represented, e.g., in the form of $\forall x, y: (x, r_s, y) \Rightarrow (x, r_t, y)$, stating that two entities linked by relation $r_s$ might also be linked by relation $r_t$. The left-hand side of the implication ``$\Rightarrow$'' is called the premise, and the right-hand side the conclusion. In this paper, we restrict $f_p\;$ to be a Horn clause rule, where the conclusion contains only a single atom and the premise is a conjunction of several atoms. The confidence level of rule $f_p$ is denoted as $\lambda_p \in [0,1]$. Rules with higher confidence levels are more likely to hold, and a confidence level of $\lambda_p=1$ indicates a hard rule which always holds with no exception. Such rules as well as their confidence levels can be extracted automatically from the KG (with the observed triple set $\mathcal{O}$ as input), by using modern rule mining systems \mbox{like AMIE and AMIE+~\cite{galarraga2013:AMIE,galarraga2015:AMIE+}.}

We then propositionalize these rules to get their groundings. Here a grounding is the logical expression with all variables instantiated with concrete entities in $\mathcal{E}$. For instance, a universally quantified rule $\forall x, y: (x, \texttt{\small BornInCountry}, y)$ $\Rightarrow (x, \texttt{\small Nationality}, y)$ could be instantiated with two entities \texttt{\small EmmanuelMacron} and \texttt{\small France}, and gives a resultant grounding $(\texttt{\small EmmanuelMacron}, \texttt{\small BornInCountry}, \texttt{\small France})$ $\Rightarrow$ $(\texttt{\small EmmanuelMacron}, \texttt{\small Nationality}, \texttt{\small France})$. Obviously, there could be a huge number of groundings, especially given a large entity vocabulary $\mathcal{E}$. In this paper, to maximize the utility for knowledge acquisition and inference, we take as valid groundings only those where premise triples are observed in $\mathcal{O}$ while conclusion triples are not. That means the aforementioned grounding will be considered as valid if the triple $(\texttt{\small EmmanuelMacron}, \texttt{\small BornInCountry}, \texttt{\small France}) \in \mathcal{O}$ but $(\texttt{\small EmmanuelMacron}, \texttt{\small Nationality}, \texttt{\small France}) \notin \mathcal{O}$. For each FOL rule $f_p$, let $\mathcal{G}_p=\{g_{pq}\}_{q=1}^{Q_p}$ denote the set of its valid groundings. All the premise triples of $g_{pq}$ are contained in $\mathcal{O}$, but the single conclusion triple is not. These conclusion triples are further used to construct our unlabeled triple set $\mathcal{U}$. That means, our unlabeled triples are those which are not directly observed in the KG but could be inferred by the rules with high probabilities.

\smallskip
\noindent\textbf{Modeling Triples and Rules.} Given the labeled triples $\mathcal{L}$, unlabeled triples $\mathcal{U}$, and the valid groundings of FOL rules $\mathcal{G}=\{\mathcal{G}_p\}_{p=1}^P$, we discuss how to model these triples and rules in the context of KG embedding. To model triples, we follow ComplEx~\cite{trouillon2016:ComplEx}, a recently proposed method which is simple and efficient while achieving state-of-the-art predictive performance. Specifically, we assume entities and relations to have complex-valued vector embeddings. Given a triple $(e_i, r_k, e_j)\in\mathcal{E}\!\times\!\mathcal{R}\!\times\!\mathcal{E}$, we score it by a multi-linear dot product:
\begin{equation}\label{eq:TripleScore}
\eta_{ijk} = \textrm{Re}(\langle \mathbf{e}_i, \mathbf{r}_k, \bar{\mathbf{e}}_j\rangle) = \textrm{Re}( \sum\nolimits_{m} [\mathbf{e}_i]_m [\mathbf{r}_k]_m [\bar{\mathbf{e}}_j]_m),
\end{equation}
where $\mathbf{e}_i,$ $\mathbf{e}_j, \mathbf{r}_k \in \mathbb{C}^d$ are the complex-valued vector embeddings associated with $e_i$, $e_j$, and $r_k$, respectively; $\bar{\mathbf{e}}_j$ is the conjugate of $\mathbf{e}_j$; $[\cdot]_m$ is the $m$-th entry of a vector; and $\textrm{Re}(\cdot)$ means taking the real part of a complex value. We further introduce a mapping function $\phi: \mathcal{E}\!\times\!\mathcal{R}\!\times\!\mathcal{E} \rightarrow (0,1)$, so as to map the score $\eta_{ijk}$ to a continuous truth value which lies in the range of $(0,1)$, i.e.,
\begin{equation}\label{eq:TripleTruthValue}
\phi(e_i, r_k, e_j) = \sigma(\eta_{ijk}) = \sigma\big(\textrm{Re}(\langle \mathbf{e}_i, \mathbf{r}_k, \bar{\mathbf{e}}_j\rangle)\big),
\end{equation}
where $\sigma(x)=1/(1+\exp(-x))$ denotes the sigmoid function. Triples with higher truth values are more likely to hold.

To model propositionalized rules (i.e. groundings), we use t-norm based fuzzy logics~\cite{hajek1998:FuzzyLogic}. The key idea is to model the truth value of a propositionalized rule as a composition of the truth values of its constituent triples, through specific logical connectives (e.g. $\wedge$ and $\Rightarrow$). For instance, the truth value of a grounded rule $(e_u, r_s, e_v) \Rightarrow (e_u, r_t, e_v)$ will be determined by the truth values of the two triples $(e_u, r_s, e_v)$ and $(e_u, r_t, e_v)$, via a composition defined by logical implication. We follow~\cite{guo2016:KALE} and define the compositions associated with logical conjunction ($\wedge$), disjunction ($\neg$), and negation ($\neg$) as:
\begin{align}
  \pi(a \wedge b) & = \pi(a) \cdot \pi(b), \\
  \pi(a \vee b)   & = \pi(a) + \pi(b) - \pi(a) \cdot \pi(b), \\
  \pi(\neg a)     & = 1 - \pi(a).
\end{align}
Here, $a$ and $b$ are two logical expressions, which can either be single triples or be constructed by combining triples with logical connectives; and $\pi(a)$ is the truth value of $a$, indicating to what degree the logical expression is true. If $a$ is a single triple, say $(e_i,r_k,e_j)$, we have $\pi(a)=\phi(e_i, r_k, e_j)$, as defined in Eq.~(\ref{eq:TripleTruthValue}). Given these compositions, the truth value of any logical expression can be calculated recursively \cite{guo2016:KALE}, e.g.,
\begin{equation}\label{eq:Implication}
\pi(a \Rightarrow b) = \pi(\neg a \vee b) = \pi(a) \cdot \pi(b) - \pi(a) + 1.
\end{equation}
Logical expressions with higher truth values have greater degrees to be true. Let $\Theta=\{\mathbf{e}\}_{e\in\mathcal{E}}\cup\{\mathbf{r}\}_{r\in\mathcal{R}}$ denote the set of all entity and relation embeddings. The proposed approach, RUGE, then aims to learn these embeddings by using the labeled triples $\mathcal{L}$, unlabeled triples $\mathcal{U}$, and valid groundings $\{\mathcal{G}_p\}_{p=1}^P$ in an iterative manner, where each iteration alternates between a soft label prediction stage and an embedding rectification stage.

\subsection{Soft Label Prediction}
This stage is to use currently learned embeddings and propositionalized rules to predict soft labels for unlabeled triples. Specifically, let $n$ be the iteration index, and $\Theta^{(n-1)}$ the set of current embeddings learned from the previous iteration. Recall that we are given a set of $P$ FOL rules with their confidence levels $\mathcal{F}=\{(f_p,\lambda_p)\}_{p=1}^{P}$, and each FOL rule $f_p$ has $Q_p$ valid groundings $\mathcal{G}_p=\{g_{pq}\}_{q=1}^{Q_p}$. Our aim is to predict a soft label $s(x_u)\in[0,1]$ for each unlabeled triple $x_u\in\mathcal{U}$, by using the current embeddings $\Theta^{(n-1)}$ and all the groundings $\mathcal{G}=\{\mathcal{G}_p\}_{p=1}^P$.

To do so, we solve a rule-constrained optimization problem, which projects truth values of unlabeled triples computed by the current embeddings into a subspace constrained by the rules. The key idea here is to find optimal soft labels that stay close to these truth values, while at the same time fitting the rules. For the first property, given each unlabeled triple $x_u\in\mathcal{U}$, we calculate its truth value $\phi(x_u)$ using the current embeddings via Eq.~(\ref{eq:TripleTruthValue}), and require the soft label $s(x_u)$ to stay close to this truth value. We measure the closeness between $s(x_u)$ and $\phi(x_u)$ with a squared loss, and try to minimize it. For the second property, we further impose rule constraints onto the soft labels $\mathcal{S}=\{s(x_u)\}$. Specifically, for each FOL rule $f_p$ and each of its groundings $g_{pq}$, we expect $g_{pq}$ to be true, i.e., $\pi(g_{pq}|\mathcal{S})\!=\!1$ with confidence $\lambda_p$. Here, $\pi(g_{pq}|\mathcal{S})$ is the conditional truth value of $g_{pq}$ given the soft labels, which can be calculated recursively with the logical compositions defined in Eq.~(3) to Eq.~(5). Take $g_{pq}:=(e_u, r_s, e_v) \Rightarrow (e_u, r_t, e_v)$ as an example, where the premise $(e_u, r_s, e_v)$ is directly observed in $\mathcal{O}$, and the conclusion $(e_u, r_t, e_v)$ is an unlabeled triple included in $\mathcal{U}$. The conditional truth value of $g_{pq}$ can then be calculated as:
\begin{equation}\label{eq:ConditionalImplication}
\pi(g_{pq}|\mathcal{S}) \!=\! \phi(e_u, r_s, e_v) \!\cdot\! s(e_u, r_t, e_v) \!-\! \phi(e_u, r_s, e_v) \!+\! 1, \!\!
\end{equation}
where $\phi(e_u, r_s, e_v)$ is a truth value defined by Eq.~(\ref{eq:TripleTruthValue}) with the current embeddings; and $s(e_u, r_t, e_v)$ is a soft label to be predicted. Comparing Eq.~(\ref{eq:ConditionalImplication}) with Eq.~(\ref{eq:Implication}), we can see that during the calculation of $\pi(g_{pq}|\mathcal{S})$, for any unlabeled triple, we use the soft label $s(\cdot)$ rather than the truth value $\phi(\cdot)$, so as to better impose rule constraints onto the soft labels $\mathcal{S}$.

Combining the two properties together and further allowing slackness for rule constraints, we finally get the following optimization problem:
\begin{align}\label{eq:SoftLabelPrediction}
\min_{\mathcal{S},\boldsymbol{\xi}}\;\; & \frac{1}{2}\sum_{x_u \in \mathcal{U}} \left(s(x_u) - \phi(x_u)\right)^2 + C\sum_{p,q} \xi_{pq}, \notag \\
\textrm{s.t.}\;\;                       & \lambda_p \left(1 - \pi(g_{pq}|\mathcal{S})\right) \leq \xi_{pq}, \; q\!=\!1,\!\cdots\!,Q_p, p\!=\!1,\!\cdots\!,P, \notag \\
                                        & \xi_{pq} \geq 0, \; q\!=\!1,\!\cdots\!,Q_p, p\!=\!1,\!\cdots\!,P, \notag \\
                                        & 0 \leq s(x_u) \leq 1, \; \forall s(x_u) \in \mathcal{S},
\end{align}
where $\xi_{pq}$ is a slack variable and $C$ the penalty coefficient. Note that confidence levels of rules (i.e. $\lambda_p$'s) are encoded in the constraints, making our approach capable of handling soft rules. Rules with higher confidence levels show less tolerance for violating the constraints. This optimization problem is convex, and can be solved efficiently with its closed-form solution:
\begin{eqnarray}\label{eq:SoftLabel}
s(x_u) = \left[ \phi(x_u) + C \sum\nolimits_{p,q} \!\!\lambda_p \nabla_{s(x_u)} \pi(g_{pq}|\mathcal{S}) \right]_0^1
\end{eqnarray}
for each $x_u \in \mathcal{U}$. Here, $\nabla_{s(x_u)} \pi(g_{pq}|\mathcal{S})$ means the gradient of $\pi(g_{pq}|\mathcal{S})$ w.r.t $s(x_u)$, which is a constant w.r.t. $\mathcal{S}$,\footnote{Note that each $g_{pq}$ contains only a single unlabeled triple, i.e., the conclusion triple. Take $\pi(g_{pq}|\mathcal{S})$ defined in Eq.~(\ref{eq:ConditionalImplication}) for example. In this case, $s(x_u)=s(e_u, r_t, e_v)$ is the soft label to be predicted and $\nabla_{s(x_u)} \pi(g_{pq}|\mathcal{S})=\phi(e_u, r_s, e_v)$ is a constant w.r.t. $\mathcal{S}$.} and $[x]_0^1=\min(\max(x,0),1)$ is a truncation function enforcing the solutions to stay within $[0,1]$. We provide the proof of convexity and detailed derivation as supplementary materials. Soft labels obtained in this way shall 1) stay close to the predictions made by the current embedding model, and 2) fit the rules as well as possible.

\subsection{Embedding Rectification}
This stage is to integrate both labeled and unlabeled triples (with hard and soft labels respectively) to update current embeddings. Specifically, we are given a set of labeled triples with their hard labels specified in $\{0,1\}$, i.e., $\mathcal{L}\!=\!\{(x_\ell,y_\ell)\}$, and also a set of unlabeled triples encoded in propositionalized rules, i.e., $\mathcal{U}=\{x_u\}$. Each unlabeled triple $x_u$ has a soft label $s(x_u)\in[0,1]$, predicted by Eq.~(\ref{eq:SoftLabel}). We would like to use these labeled and unlabeled triples to learn the updated embeddings $\Theta^{(n)}$. Here $n$ is the iteration index.

To this end, we minimize a global loss over $\mathcal{L}$ and $\mathcal{U}$, so as to find embeddings which can predict the true hard labels for triples contained in $\mathcal{L}$, while imitating the soft labels for those contained in $\mathcal{U}$. The optimization problem is:
\begin{align}\label{eq:EmbeddingRectification}
\min_{\Theta} \frac{1}{|\mathcal{L}|} \sum_{\mathcal{L}} \ell (\phi(x_\ell), y_\ell) + \frac{1}{|\mathcal{U}|}\sum_{\mathcal{U}} \ell (\phi(x_u), s(x_u)),
\end{align}
where $\ell(x,y)=-y\log x - (1-y)\log(1-x)$ is the cross entropy; and $\phi(\cdot)$ is a function w.r.t. $\Theta$ defined by Eq.~(\ref{eq:TripleTruthValue}). We further impose $L_2$ regularization on the parameters $\Theta$ to avoid overfitting. Gradient descent algorithms can be used to solve this problem. Embeddings learned in this way will 1) be compatible with all the labeled triples, and 2) absorb rule knowledge carried by the unlabeled triples.

\subsection{Whole Procedure}
Algorithm~\ref{alg:RUGE} summarizes the iterative learning procedure of our approach. To enable efficient learning, we use an online scheme in mini-batch mode. At each iteration, we sample a mini-batch $\mathcal{L}_b$, $\mathcal{U}_b$, and $\mathcal{G}_b$ from the labeled triples $\mathcal{L}$, unlabeled triples $\mathcal{U}$, and propositionalized rules $\mathcal{G}$, respectively (line 3).\footnote{We first sample $\mathcal{L}_b$ from $\mathcal{L}$. $\mathcal{G}_b$ is then constructed by those whose premise triples are all contained in $\mathcal{L}_b$ but conclusion triples are not. These conclusion triples are further used to construct $\mathcal{U}_b$.} Soft label prediction and embedding rectification are then conducted locally on these mini-batches (line 4 and line 5 respectively). This iterative procedure captures the interactive nature between embedding learning and logical inference: given current embeddings, logic rules can be used to perform approximate inference and predict soft labels for unlabeled triples; these newly labeled triples carry rich rule knowledge and will in turn help to learn better embeddings. In this way, knowledge contained in logic rules can be fully transferred into the learned embeddings. Note also that our approach is flexible enough to handle soft rules with various confidence levels extracted automatically from the KG.

\subsection{Discussions}
We further analyze the space and time complexity, and discuss possible extensions of our approach.

\smallskip
\noindent\textbf{Complexity.} RUGE follows ComplEx to represent entities and relations as complex-valued vectors, hence has a space complexity of $O(n_e d + n_r d)$ which scales linearly w.r.t. $n_e$, $n_r$, and $d$. Here, $n_e$ is the number of entities, $n_r$ the number of relations, and $d$ the dimensionality of the embedding space. During the learning procedure, each iteration requires a time complexity of $O(\tau(n_\ell d + n_u d))$, where $n_\ell$/$n_u$ is the average number of labeled/unlabeled triples in a mini-batch, and $\tau$ the number of inner epochs used for embedding rectification (cf. Eq.~(\ref{eq:EmbeddingRectification})). In practice, we usually have $n_u \ll n_\ell$ (see Table~\ref{tab:Dataset} for the number of labeled and unlabeled triples used on our datasets), and we can also set $\tau$ to a very small value, e.g., $\tau=1$. That means, RUGE has almost the same time complexity as those most efficient KG embedding techniques (e.g. ComplEx) which require $O(n_\ell d)$ per iteration during training.\footnote{Such techniques often use SGD in mini-batch mode for training, and sample a mini-batch of $n_\ell$ labeled triples at each iteration.} In addition, RUGE further requires preprocessing steps before training, i.e., rule mining and propositionalization. But these steps are performed only once, and not required during the iterations.

\smallskip
\noindent\textbf{Extensions.} Our approach is quite generic and flexible. 1) The idea of iteratively injecting logic rules can be applied to enhance a wide variety of embedding models, as long as an appropriate scoring function is accordingly designed, e.g., the one defined in Eq.~(\ref{eq:TripleScore}) by ComplEx. 2) Various types of rules can be incorporated as long as they can be modeled by the logical compositions defined in Eq.~(3) to Eq.~(5), and we can even use other types of t-norm fuzzy logics to define such compositions. 3) Rules with different confidence levels can be handled in a unified manner.

\begin{algorithm}[t]
\small
\caption{Iterative Learning Procedure of RUGE}\label{alg:RUGE}
\begin{algorithmic}[1]
\REQUIRE {Labeled triples $\mathcal{L} = \{(x_\ell,y_\ell)\}$ \\
$\quad\quad\;$ Unlabeled triples $\mathcal{U}=\{x_u\}$ \\
$\quad\quad\;$ FOL rules $\mathcal{F}\!\!=\!\!\{(f_p,\lambda_p)\}$ and their groundings $\mathcal{G}\!\!=\!\!\{g_{pq}\}$}
\STATE Randomly initialize entity and relation embeddings $\Theta^{(0)}$
\FOR {$n = 1 : N$}
    \STATE Sample a mini-batch $\mathcal{L}_b$ / $\mathcal{U}_b$ / $\mathcal{G}_b$ from $\mathcal{L}$ / $\mathcal{U}$ / $\mathcal{G}$
    \STATE {$\mathcal{S}_b \!\gets\! \textrm{SoftLabelPrediction}\,(\mathcal{U}_b, \mathcal{G}_b, \Theta^{(n\!-\!1)})$  $\quad\;\;\triangleright$ cf. Eq.~(\ref{eq:SoftLabel})}
    \STATE {$\Theta^{(n)} \!\gets\! \textrm{EmbeddingRectification}\,(\mathcal{L}_b, \mathcal{U}_b, \mathcal{S}_b)$ $\;\triangleright$ cf. Eq.~(\ref{eq:EmbeddingRectification})}
\ENDFOR
\ENSURE {$\Theta^{(N)}$}
\end{algorithmic}
\end{algorithm}

\section{Experiments}
We evaluate RUGE in the link prediction task. This task is to complete a triple $(e_i, r_k, e_j)$ with $e_i$ or $e_j$ missing, i.e., to predict $e_i$ given $(r_k, e_j)$ or $e_j$ given $(e_i, r_k)$.

\smallskip
\noindent\textbf{Datasets.} We use two datasets: FB15K and YAGO37. The former is a subgraph of Freebase containing 1,345 relations and 14,951 entities, released by Bordes et al.~\shortcite{bordes2013:TransE}.\footnote{https://everest.hds.utc.fr/doku.php?id=en:smemlj12} The latter is extracted from the core facts of YAGO3.\footnote{http://www.mpi-inf.mpg.de/departments/databases-and-information-systems/research/yago-naga/yago/downloads/} During the extraction, entities appearing less than 10 times are discarded. The final dataset consists of 37 relations and 123,189 entities. Triples on both datasets are split into training, validation, and test sets, used for model training, hyperparameter tuning, and evaluation, respectively. We use the original split for FB15K, and draw a split of 989,132/50,000/50,000 triples for YAGO37.

Note that on both datasets, the training sets contain only positive triples. Negative triples are generated using the local closed world assumption~\cite{dong2014:KnowledgeVault}. This negative sampling procedure is performed at runtime for each batch of training positive triples. Such positive and negative triples (along with their hard labels) form our \textit{labeled triple set}.

We further employ AMIE+~\cite{galarraga2015:AMIE+}\footnote{https://www.mpi-inf.mpg.de/departments/databases-and-information-systems/research/yago-naga/amie/} to automatically extract Horn clause rules from each dataset, with the \textit{training} set as input. To enable efficient extraction, we consider rules with length not longer than 2 and confidence levels not less than 0.8.\footnote{AMIE+ provides two types of confidence, i.e. standard confidence and PCA confidence. This paper uses PCA confidence.} The length of a Horn clause rule is the number of atoms appearing in its premise, e.g., $\forall x, y:$ $(x, \texttt{\small BornInCountry}, y) \Rightarrow (x, \texttt{\small Nationality}, y)$ has the length of 1. And the confidence threshold of 0.8 leads to the best performance on both datasets (detailed later). Using this setting, we extract 454 (universally quantified) Horn clause rules from FB15K, and 16 such rules from YAGO37. Table~\ref{tab:Rules} shows some examples with their confidence levels.

\begin{table}[t]
    \centering\scriptsize
    \caption{\label{tab:Rules} Horn clause rules with confidence levels extracted by AMIE+ from FB15K (top) and YAGO37 (bottom).}
    \begin{tabular*}{0.47 \textwidth}{@{\extracolsep{\fill}}@{}lc@{}}
        \toprule
        $\textrm{/location/people\_born\_here}(\!\!\:x\!\!\;,\!y\!\!\:) \!\!\!\;\Rightarrow\!\!\!\; \textrm{/people/place\_of\_birth}(\!\!\:y\!\!\;,\!x\!\!\:)$                    & 1.00 \\
        $\textrm{/director/film}(\!\!\:x\!\!\;,\!y\!\!\:) \!\!\!\;\Rightarrow\!\!\!\; \textrm{/film/directed\_by}(\!\!\:y\!\!\;,\!x\!\!\:)$                                        & 0.99 \\
        $\textrm{/film/directed\_by}(\!\!\:x\!\!\;,\!y\!\!\:) \!\!\!\;\wedge\!\!\!\; \textrm{/person/language}(\!\!\:y\!\!\;,\!z\!\!\:) \!\!\!\;\Rightarrow\!\!\!\; \textrm{/film/language}(\!\!\:x\!\!\;,\!z\!\!\:)$ & 0.88 \\
        \midrule
        $\textrm{isMarriedTo}(\!\!\:x\!\!\;,\!y\!\!\:) \!\!\!\;\Rightarrow\!\!\!\; \textrm{isMarriedTo}(\!\!\:y\!\!\;,\!x\!\!\:)$                                                  & 0.97 \\
        $\textrm{hasChild}(\!\!\:x\!\!\;,\!y\!\!\:) \!\!\!\;\wedge\!\!\!\; \textrm{isCitizenOf}(\!\!\:y\!\!\;,\!z\!\!\:) \!\!\!\;\Rightarrow\!\!\!\; \textrm{isCitizenOf}(\!\!\:x\!\!\;,\!z\!\!\:)$    & 0.94 \\
        $\textrm{playsFor}(\!\!\:x\!\!\;,\!y\!\!\:) \!\!\!\;\Rightarrow\!\!\!\; \textrm{isAffiliatedTo}(\!\!\:x\!\!\;,\!y\!\!\:)$                                                  & 0.86 \\
        \bottomrule
    \end{tabular*}
\end{table}

Then, we instantiate these rules with concrete entities, i.e., propositionalization. Propositionalized rules whose premise triples are all contained in the \textit{training} set (while conclusion triples are not) are taken as \textit{valid groundings} and used during embedding learning. We obtain 96,724 valid groundings on FB15K and 72,670 on YAGO37. Conclusion triples of these valid groundings are further collected to form our \textit{unlabeled triple set}. We finally get 74,707 unlabeled triples on FB15K and 69,680 on YAGO37. Table~\ref{tab:Dataset} provides some statistics of the two datasets.

\begin{table}[t]
    \centering\scriptsize\setlength{\tabcolsep}{2pt}
    \caption{\label{tab:Dataset} Statistics of datasets, where $n_e$/$n_r$ denotes the number of entities/relations, $n_\ell$/$n_u$/$n_g$ is the number of labeled triples/unlabeled triples/valid groundings used for training, and $n_v$/$n_t$ denotes the number of validation/test triples.}
    \begin{tabular*}{0.47 \textwidth}{@{\extracolsep{\fill}}@{}lrrrrrrr@{}}
        \toprule
        & & & \multicolumn{3}{c}{Train} & \multicolumn{1}{c}{Valid} & \multicolumn{1}{c}{Test} \\\cmidrule{4-6}\cmidrule{7-7}\cmidrule{8-8}
        Dataset & \multicolumn{1}{c}{$n_e$} & \multicolumn{1}{c}{$n_r$} & \multicolumn{1}{c}{$n_\ell$} & \multicolumn{1}{c}{$n_u$} & \multicolumn{1}{c}{$n_g$} & \multicolumn{1}{c}{$n_v$} & \multicolumn{1}{c}{$n_t$} \\
        \midrule
        FB15K  & 14,951    & 1,345      & 483,142   & 74,707  & 96,724  & 50,000  & 59,071 \\
        YAGO37 & 123,189   & 37         & 989,132   & 69,680  & 72,670  & 50,000  & 50,000 \\
        \bottomrule
    \end{tabular*}
\end{table}

\smallskip
\noindent\textbf{Evaluation Protocol.} To evaluate the performance in link prediction, we follow the standard protocol used in~\cite{bordes2013:TransE}. For each test triple $(e_i, r_k, e_j)$, we replace the head entity $e_i$ with each entity $e_i'\!\in\!\mathcal{E}$, and calculate the score for $(e_i', r_k, e_j)$. Ranking these scores in descending order, we get the rank of the correct entity $e_i$. Similarly, we can get another rank by replacing the tail entity. Aggregated over all test triples, we report three metrics: 1) the mean reciprocal rank (MRR), 2) the median of the ranks (MED), and 3) the proportion of ranks no larger than $n$ (HITS@N). During this ranking process, we remove corrupted triples which already exist in either the training, validation, or test set, since they themselves are true triples. This corresponds to the ``filtered'' setting in~\cite{bordes2013:TransE}.

\smallskip
\noindent\textbf{Comparison Settings.} We compare RUGE with four state-of-the-art basic embedding models, including TransE~\cite{bordes2013:TransE}, DistMult~\cite{yang2015:DistMult}, HolE~\cite{nickel2016:HolE}, and ComplEx~\cite{trouillon2016:ComplEx}. These basic models rely only on triples observed in a KG and use no rules. We further take PTransE~\cite{lin2015:PTransE} and KALE~\cite{guo2016:KALE} as additional baselines. Both of them are extensions of TransE, with the former integrating relation paths (Horn clauses), and the latter FOL rules (hard rules) in a one-time injection manner. In contrast, RUGE incorporates soft rules and transfers rule knowledge into KG embedding in an iterative manner.

We use the code provided by Trouillon et al.~\shortcite{trouillon2016:ComplEx}\footnote{https://github.com/ttrouill/complex} for TransE, DistMult, and ComplEx, and reimplement HolE so that all these four basic models share the identical mode of optimization, i.e., SGD with AdaGrad~\cite{duchi2011:AdaGrad} and gradient normalization. As such, we reproduce the results of TransE, DistMult, and ComplEx reported on FB15K~\cite{trouillon2016:ComplEx}, and improve the results of HolE substantially compared to those reported in the original paper~\cite{nickel2016:HolE}.\footnote{HolE in its original implementation uses SGD with AdaGrad, but no gradient normalization.} The code for PTransE is provided by its authors.\footnote{https://github.com/thunlp/KB2E} We implement KALE and RUGE in Java, both using SGD with AdaGrad and gradient normalization to facilitate a fair comparison.

There are two types of loss functions that could be used for these baselines, i.e., the logistic loss or the pairwise ranking loss~\cite{nickel2016:HolE}. Trouillon et al. \shortcite{trouillon2016:ComplEx} have recently demonstrated that the logistic loss generally performs better than the pairwise ranking loss, except for TransE. So, for TransE and its extensions (PTransE and KALE) we use the pairwise ranking loss, and for all the other baselines we use the logistic loss. To extract relation paths for PTransE, we follow the optimal configuration reported in \cite{lin2015:PTransE}, where paths constituted by at most 3 relations are included. For KALE and RUGE, we use the same set of propositionalized rules to make it a fair comparison.\footnote{KALE takes all these groundings as hard rules. This approximation works quite well with an appropriate confidence threshold.}

For all the methods, we create 100 mini-batches on each dataset, and tune the embedding dimensionality $d$ in $\{50,$ $100, 150, 200\}$, the number of negatives per positive triple $\alpha$ in $\left\{ 1, 2, 5, 10\right\}$, the initial learning rate $\gamma$ in $\{0.01, 0.05, 0.1,$ $0.5, 1.0\}$, and the $L_2$ regularization coefficient $\lambda$ in $\{0.001,$ $0.003, 0.01, 0.03, 0.1\}$. For TransE and its extensions which use the pairwise ranking loss, we further tune the margin $\delta$ in $\{0.1, 0.2, 0.5, 1, 2, 5, 10\}$. The slackness penalty $C$ in RUGE (cf. Eq.~(\ref{eq:SoftLabelPrediction})) is selected from $\{0.001, 0.01, 0.1, 1\}$, and the number of inner iterations (cf. Eq.~(\ref{eq:EmbeddingRectification})) is fixed to $\tau=1$. Best models are selected by early stopping on the validation set (monitoring MRR), with at most 1000 iterations over the training set. The optimal configurations for RUGE are: $d=$ $200$, $\alpha=10$, $\gamma=0.5$, $\lambda=0.01$, $C=0.01$ on FB15K; and $d=150$, $\alpha=10$, $\gamma\!=\!1.0$, $\lambda\!=\!0.003$, $C\!=\!0.01$ on YAGO37.

\begin{table*}[t]
    \centering\footnotesize\setlength{\tabcolsep}{5pt}
    \caption{\label{tab:LinkPrediction} Link prediction results on the test sets of FB15K and YAGO37. As baselines, rows 1-4 are the four basic models which use triples alone, and rows 5-6 further integrate logic rules (or relation paths) in a one-time injection manner.}
    \begin{tabular*}{1 \textwidth}{@{\extracolsep{\fill}}@{}lccccccccccccc@{}}
    \toprule
    & \multicolumn{6}{c}{FB15K} && \multicolumn{6}{c}{YAGO37} \\\cmidrule{2-7}\cmidrule{9-14}
    & & & \multicolumn{4}{c}{HITS@N} & & & & \multicolumn{4}{c}{HITS@N} \\\cmidrule{4-7}\cmidrule{11-14}
    Method  & MRR & MED      & 1    & 3    & 5    & 10   && MRR  & MED & 1 & 3 & 5 & 10 \\
    \midrule
    TransE  &0.400 &4.0      &0.246 &0.495 &0.576 &0.662 &&0.303 &13.0     &0.218 &0.336 &0.387 &0.475 \\
    DistMult&0.644 &\bf{1.0} &0.532 &0.730 &0.769 &0.812 &&0.365 &6.0      &0.262 &0.411 &0.493 &0.575 \\
    HolE    &0.600 &2.0      &0.485 &0.673 &0.722 &0.779 &&0.380 &7.0      &0.288 &0.420 &0.479 &0.551 \\
    ComplEx &0.690 &\bf{1.0} &0.598 &0.756 &0.793 &0.837 &&0.417 &\bf{4.0} &0.320 &0.471 &0.533 &\bf{0.603} \\
    PTransE &0.679 &\bf{1.0} &0.565 &0.768 &0.810 &0.855 &&0.403 &9.0      &0.339 &0.444 &0.473 &0.506 \\
    KALE    &0.523 &2.0      &0.383 &0.616 &0.683 &0.762 &&0.321 &9.0      &0.215 &0.372 &0.438 &0.522 \\
    \midrule
    RUGE    &\bf{0.768}&\bf{1.0}&\bf{0.703}&\bf{0.815}&\bf{0.836}&\bf{0.865}&
            &\bf{0.431}&\bf{4.0}&\bf{0.340}&\bf{0.482}&\bf{0.541}&\bf{0.603}\\
    \bottomrule
    \end{tabular*}
\end{table*}

\smallskip
\noindent\textbf{Link Prediction Results.} Table~\ref{tab:LinkPrediction} shows the results of these methods on the test sets of FB15K and YAGO37. The results indicate that RUGE significantly and consistently outperforms all the baselines on both datasets and in all metrics. It beats not only the four basic models which use triples alone (TransE, DistMult, HolE, and ComplEx), but also PTransE and KALE which further incorporate logic rules (or relation paths) in a one-time injection manner. This demonstrates the superiority of injecting logic rules into KG embedding, particularly in an iterative manner. Compared to the best performing baseline ComplEx (this is also the model based on which RUGE is designed), RUGE achieves an improvement of 11\%/18\% in MRR/HITS@1 on FB15K, and an improvement of 3\%/6\% on YAGO37. The improvements on FB15K are more substantial than those on YAGO37. The reason is probably that FB15K contains more relations from which a good range of rules can be extracted (454 universally quantified rules from FB15K, and 16 from YAGO37).

\begin{figure}[t]
    \centering
    \label{fig:Confidence-FB15K}
        \includegraphics[width=0.367\textwidth]{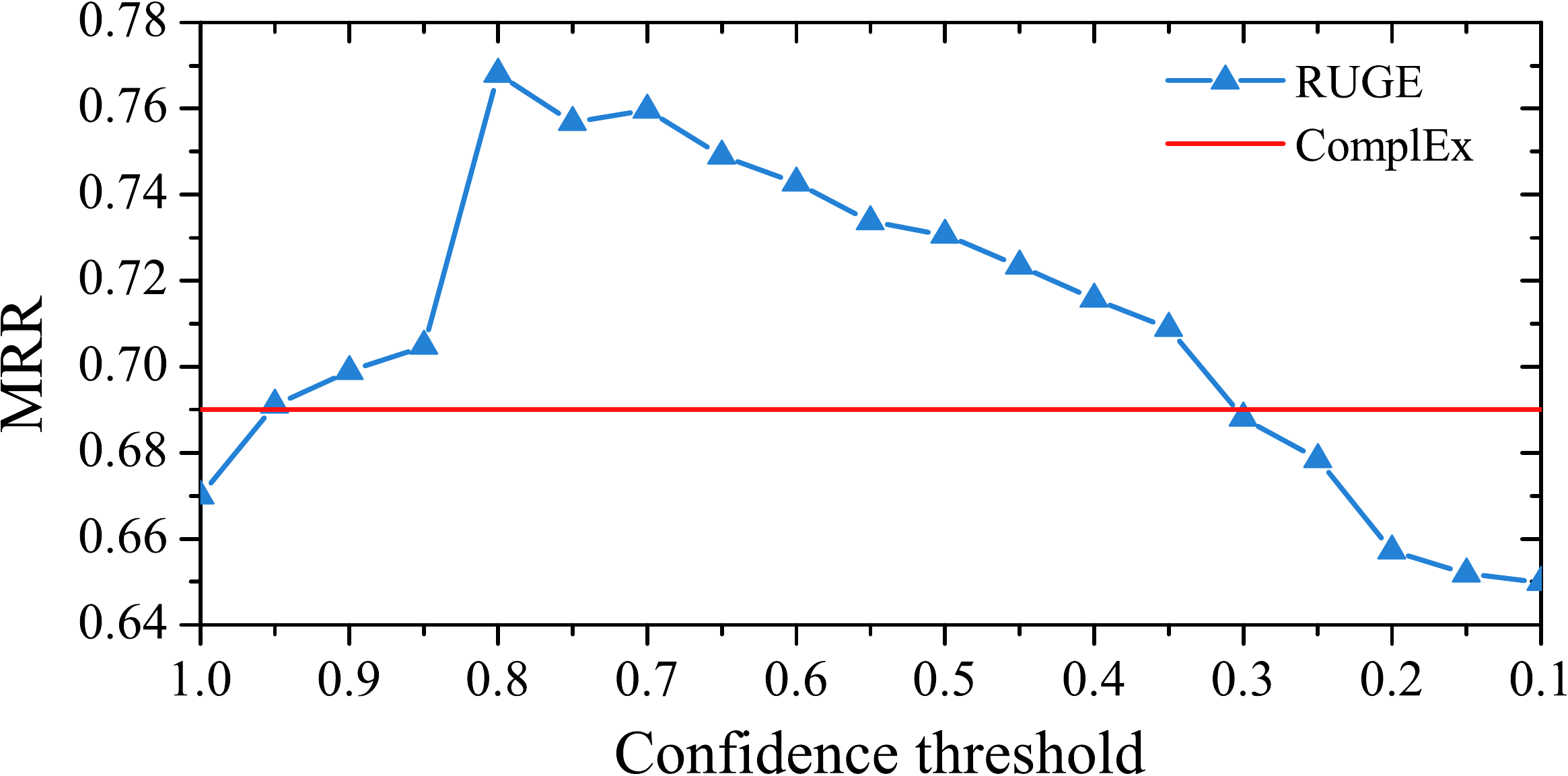}
  \caption{MRR achieved by RUGE with different confidence thresholds on the test set of FB15K.}\label{fig:Confidence}
\end{figure}

\smallskip
\noindent\textbf{Influence of Confidence Levels.} We further investigate the influence of the threshold of rules' confidence levels used in RUGE. To do so, we fix all the hyperparameters to the optimal configurations determined by the previous experiment, and vary the confidence threshold in $[0.1,1]$ with a step 0.05. Fig.~\ref{fig:Confidence} shows MRR achieved by RUGE with various thresholds on the test set of FB15K. We can see that the threshold of 0.8 is a good tradeoff and indeed performs best. A threshold higher than that will reduce the number of rules that can be extracted, while a one lower than that might introduce too many less credible rules. Both hurt the performance. However, even so, RUGE outperforms ComplEx by a large margin, with the threshold set in a broad range of $[0.35, 0.9]$. This observation indicates that soft rules, even those with moderate confidence levels, are highly beneficial to KG embedding despite their uncertainties. 

\begin{table}[t]
    \centering\footnotesize\setlength{\tabcolsep}{2pt}
    \caption{\label{tab:Runtime} Runtime (in sec.) on FB15K and YAGO37. Extr. is the time required for rule/path extraction, Prop. for propositionalization, and Lean. for training per iteration.}
    \begin{tabular*}{0.47 \textwidth}{@{\extracolsep{\fill}}@{}lrrrrrrr@{}}
    \toprule
    & \multicolumn{3}{c}{FB15K} && \multicolumn{3}{c}{YAGO37} \\\cmidrule{2-4}\cmidrule{6-8}
    Method  & Extr. & Prop. & Learn. && Extr. & Prop. & Learn. \\
    \midrule
    ComplEx & ---   & --- & 11.4 && ---     & ---   & 49.5 \\
    PTransE & 868.4 & --- & 46.5 && 13939.5 & ---   & 23.8 \\
    KALE    & 43.1  & 4.0 & 4.8  && 337.5   & 13.8  & 27.5 \\
    RUGE    & 43.1  & 4.0 & 14.1 && 337.5   & 13.8  & 55.2 \\
    \bottomrule
    \end{tabular*}
\end{table}

\smallskip
\noindent\textbf{Comparison of Runtime.} Finally, we compare RUGE with ComplEx, PTransE, and KALE in their runtime.\footnote{The other three baselines are implemented in Python and much slower. So they are not considered here.} ComplEx is a basic model which only requires model training. RUGE as well as the other two baselines further require preprocessing of rule/path extraction and propositionalization. Table~\ref{tab:Runtime} lists the runtime of these methods required for each step on FB15K and YAGO37. Here, to facilitate a fair comparison, we set $d\!=\!200$ (embedding dimensionality) and $\alpha\!=\!2$ (number of negatives per positive triple) for all the methods. Other hyperparameters are fixed to their optimal configurations determined in link prediction. We can see that RUGE is still quite efficient despite integrating additional rules. The average training time per iteration increases from 11.4 to 14.1 on FB15K, and from 49.5 to 55.2 on YAGO37. The preprocessing steps, although performed only once, are also highly efficient, requiring much less time compared to PTransE.

\section{Conclusion}
This paper proposes a novel paradigm that learns entity and relation embeddings with iterative guidance from soft rules, referred to as RUGE. It enables an embedding model to learn simultaneously from labeled triples, unlabeled triples, and soft rules in an iterative manner. Each iteration alternates between 1) a soft label prediction stage which predicts soft labels for unlabeled triples using currently learned embeddings and soft rules, and 2) an embedding rectification stage which further integrates both labeled and unlabeled triples to update current embeddings. This iterative procedure may better transfer the knowledge contained in logic rules into the learned embeddings. Link prediction results on Freebase and YAGO show that RUGE achieves significant and consistent improvements over state-of-the-art baselines. Moreover, RUGE demonstrates the usefulness of automatically extracted soft rules. Even those with moderate confidence levels can be highly beneficial to KG embedding.

\section*{Acknowledgments}
The authors would like to thank all the reviewers for their insightful and valuable suggestions, which significantly improve the quality of this paper. This work is supported by the National Key Research and Development Program of China (grants No. 2016YFB0801003 and No. 2016QY03D0503), the Fundamental Theory and Cutting Edge Technology Research Program of the Institute of Information Engineering, Chinese Academy of Sciences (grant No. Y7Z0261101), and the National Natural Science Foundation of China (grant No. 61402465).


\clearpage
\includepdfmerge{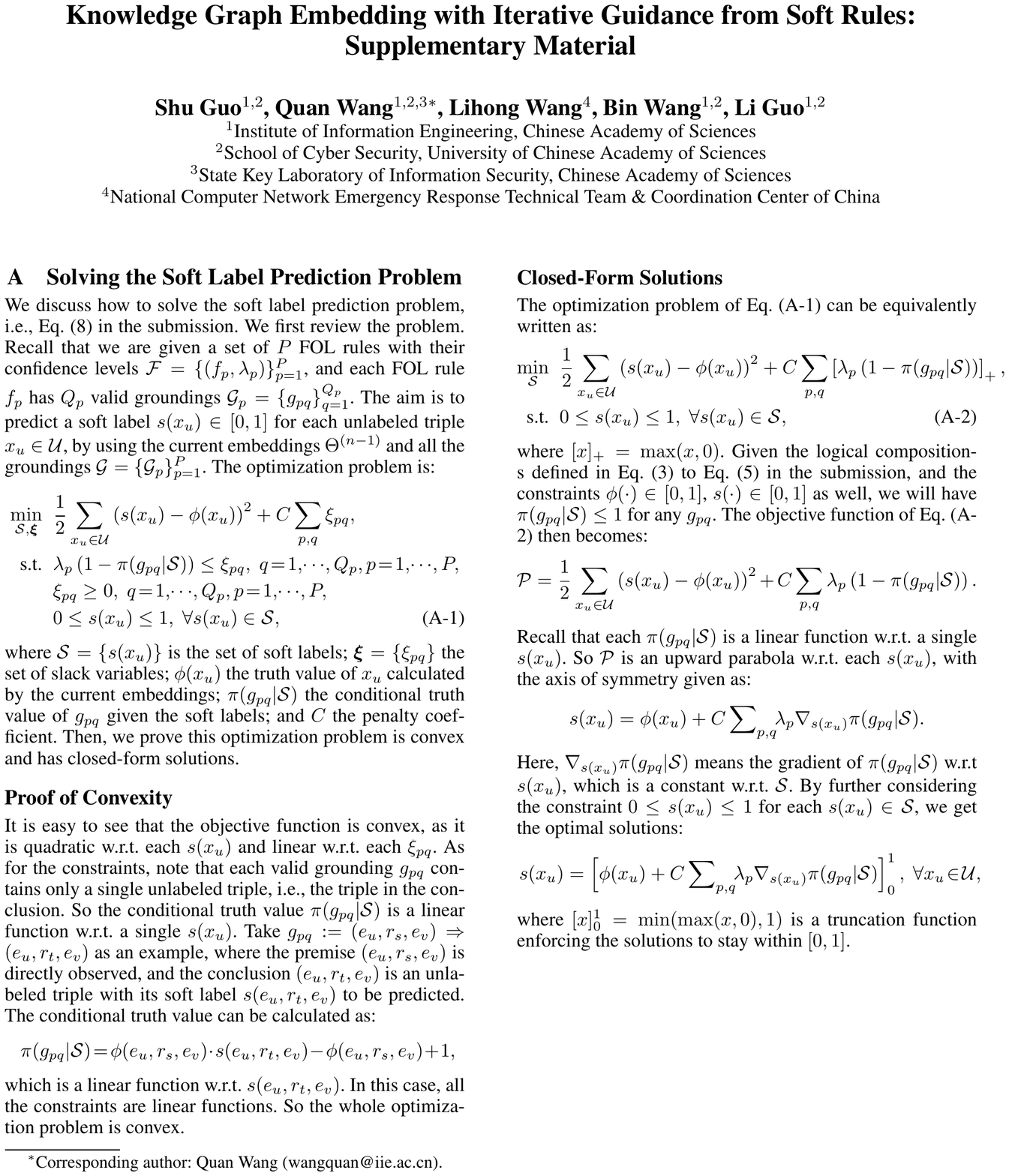}
\end{document}